\documentclass[conference]{IEEEtran}
\IEEEoverridecommandlockouts
\usepackage[justification=justified]{caption}
\usepackage{amsmath}
\usepackage{amsfonts}
\usepackage{graphicx}
\usepackage{textcomp}
\usepackage{xcolor}
\usepackage[numbers]{natbib}
\usepackage{url}
\usepackage{relsize}
\usepackage[utf8]{inputenc} 
\usepackage[T1]{fontenc}    
\usepackage{url}            
\usepackage{booktabs}       
\usepackage{amsfonts}       
\usepackage{nicefrac}       
\usepackage{microtype}      
\usepackage{xcolor}         
\usepackage{color, colortbl}
\definecolor{LightCyan}{rgb}{0.88,1,1}
\definecolor{deepskyblue}{rgb}{0, 191, 255}
\definecolor{fuchsia}{rgb}{100, 0, 100}
\usepackage{enumerate}
\usepackage{enumitem}
\usepackage{bbm}
\usepackage{wrapfig}
\usepackage{balance}
\usepackage{subcaption}
\usepackage{booktabs}
\usepackage{multirow}
\usepackage{algorithm}
\usepackage{algpseudocode}
\usepackage{xcolor}

\usepackage{amsmath}

\newtheorem{proposition}{Proposition}
\newtheorem{theorem}{Theorem}

\DeclareMathOperator*{\argmax}{arg\,max}

\usepackage{cases}
\usepackage{xcolor}
\usepackage{amssymb}
\usepackage{tcolorbox}
\usepackage{graphicx}
\usepackage{textcomp}
\usepackage{multirow}
\usepackage{float}
\usepackage{tikz}
\newcommand{\cdotsTwo}{%
  \mathinner{{\cdotp}{\cdotp}}%
}

\definecolor{myred}{RGB}{255,0,0}
\usepackage{hyperref}
\usepackage{cleveref}
\hypersetup{
    colorlinks=false,
    linkcolor=black,
    filecolor=black,      
    urlcolor=black,
    pdftitle={Overleaf Example},
    pdfpagemode=FullScreen,
}
\newcommand\copyrighttext{%
 \footnotesize\textcopyright © 2025 IEEE. Personal use of this material is permitted. Permission from IEEE must be obtained for all other uses, in any current or future media, including reprinting/republishing this material for advertising or promotional purposes, creating new collective works, for resale or redistribution to servers or lists, or reuse of any copyrighted component of this work in other works.}
\newcommand\copyrightnotice{%
\begin{tikzpicture}[remember picture,overlay]
\node[anchor=south,yshift=30pt] at (current page.south) {\fbox{\parbox{\dimexpr\textwidth-\fboxsep-\fboxrule\relax}{\copyrighttext}}};
\end{tikzpicture}%
}

\begin{document}
\title{
Optimized Local Updates in Federated Learning via Reinforcement Learning \\
\vspace{2mm}
\footnotesize{*Note: Accepted at IEEE International Joint Conference on Neural Networks (IJCNN) 2025.}
\vspace{-5mm}
}
\author{\IEEEauthorblockN{Ali Murad}
\IEEEauthorblockA{
\textit{Auburn University}\\
Auburn AL, USA \\
azm0269@auburn.edu}
\and
\IEEEauthorblockN{Bo Hui}
\IEEEauthorblockA{
\textit{Auburn University}\\
Auburn AL, USA \\
bzh0055@auburn.edu}
\and
\IEEEauthorblockN{Wei-Shinn Ku}
\IEEEauthorblockA{
\textit{Auburn University}\\
Auburn AL, USA \\
wzk0004@auburn.edu}
}
\maketitle
\copyrightnotice
\begin{abstract}
Federated Learning (FL) is a distributed framework for collaborative model training over large-scale distributed data, enabling higher performance while maintaining client data privacy. However, the nature of model aggregation at the centralized server can result in a performance drop in the presence of non-IID data across different clients. We remark that training a client locally on more data than necessary does not benefit the overall performance of all clients. In this paper, we devise a novel framework that leverages a Deep Reinforcement Learning (DRL) agent to select an optimized amount of data necessary to train a client model without oversharing information with the server. Starting without awareness of the client's performance, the DRL agent utilizes the change in training loss as a reward signal and learns to optimize the amount of training data necessary for improving the client's performance. Specifically, after each aggregation round, the DRL algorithm considers the local performance as the current state and outputs the optimized weights for each class, in the training data, to be used during the next round of local training. In doing so, the agent learns a policy that creates an optimized partition of the local training dataset during the FL rounds. After FL, the client utilizes the entire local training dataset to further enhance its performance on its own data distribution, mitigating the non-IID effects of aggregation. Through extensive experiments, we demonstrate that training FL clients through our algorithm results in superior performance on multiple benchmark datasets and FL frameworks. Our code is available at \color{blue}{\url{https://github.com/amuraddd/optimized_client_training.git}}.
\end{abstract}
\vspace{-0.3cm}
\section{Introduction}
The growth in computing power has enabled learning algorithms to learn from increasingly more data. In general, learning from more data has been assumed to lead to higher performance~\cite{goodfellow2016deep}. However, the amount of data required by the learning algorithm remains an arbitrary choice driven by personal preference and past experience. Meanwhile, in distributed systems, the use of more data for model training can pose privacy risks, particularly in settings where data can be leaked or used for personal identification~\cite{pmlr-v202-allouah23a}~\cite{DBLP:conf/iclr/0050BSCZ0S24}. Federated Learning (FL) has emerged as a powerful framework for distributed learning through which multiple parties, also known as clients, collaborate to train global models without sharing their data~\cite{li2021survey}~\cite{ mcmahan2017communication}. In centralized FL clients perform limited training on local datasets while a centralized server aggregates the client parameters using various aggregation methods. In this way, the data of each client is kept private, and superior performance can be achieved.

\begin{figure*}[t]
    \centering
    \begin{subfigure}[t]{0.45\textwidth}
        \centering
        \includegraphics[width=0.75\linewidth, trim=0 0 0 0, clip]{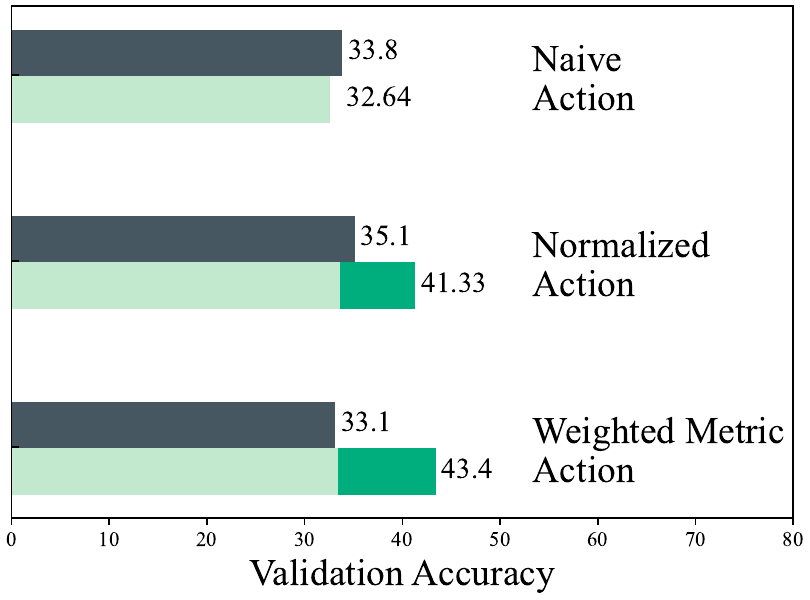}
        \caption{Naive vs. Optimized Accuracy. The y-axis shows RL agent actions. Each action is described in detail in the method section.}
        \label{fig:val_acc_naive_optimal}
    \end{subfigure}
    \hspace{1mm}
    \begin{subfigure}[t]{0.45\textwidth}
        \centering
        \includegraphics[width=0.75\linewidth, trim=0 -30 0 0, clip]{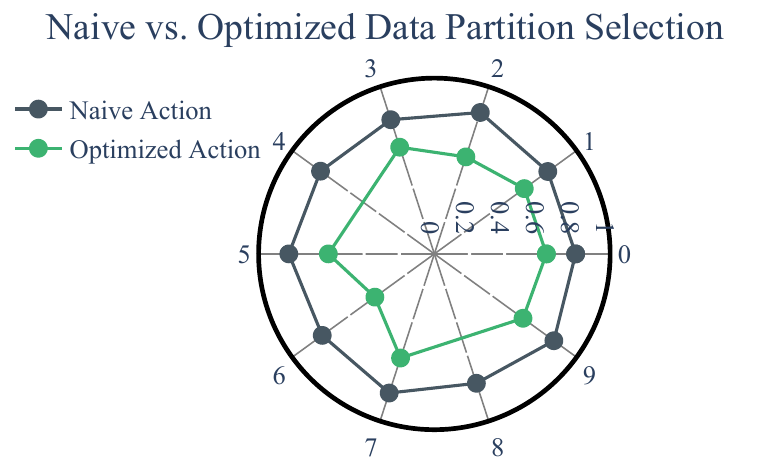}
        \caption{Data selection in each of 10 classes of CIFAR-10. The radii of the circle show the proportion of data selected for each class.}
        \label{fig:data_selection_naive_optimal}
    \end{subfigure}
    \caption{Naive vs. Optimized clients.}
    \label{fig:motivation}
\end{figure*}

Our main motivation is that training a client locally on more data than necessary does not benefit the overall performance of all clients. The motivation has been empirically verified in Figure~\ref{fig:val_acc_naive_optimal} where using sub-dataset (green color) outperforms using all data (black color). This is because the data across different clients are not independent, and two sets of data can cancel out their effects on the model update after aggregation. Finding the optimized amount of data necessary for local training enables the client to optimize its own performance while maximizing contributions to the global model through aggregation. Moreover, we empirically find that at the end of the FL rounds, the client benefits from unused data in the prior learning rounds by training the final aggregated parameters on the complete local training dataset. This unused data provides the client with fresh information that enriches the model parameters.

In this paper, we build a novel FL framework to find the optimized partition of local training data. We first introduce the notion of an optimized client, which finds the optimized ratio of local training data to train the local model without oversharing local information with the server. To maintain a distinction between the optimized clients and all other clients in the FL scheme, we refer to the remaining clients as naive clients. The selection of the optimized partition is demonstrated in Figure~\ref{fig:data_selection_naive_optimal} where the radii of the unit circle represent the proportion of data used in naive clients and an optimized client during FL. 
As shown in Figure~\ref{fig:val_acc_naive_optimal}, using an optimized partition of training data, to train the optimized client, does not impact the performance of naive clients. Moreover, our algorithm can improve the performance of the optimized client compared with the original FL strategy. To build the optimized client, we introduce Deep Reinforcement Learning (DRL) during the FL rounds to train an agent. The DRL agent takes model performance on the client's local dataset in each FL round as the current state. An action is defined as changing the optimized ratio of training data to be used for local training. The optimized client treats the FL setting as the environment and the reward for the DRL agent is the reduction in training loss. The action taken by the agent selects a partition of local data for each class in the dataset. The selected partition is then used by the optimized client for local training to optimize a given metric (i.e., \text{F1}-Score, Recall, Precision, Accuracy, etc.) for each class in the dataset. As FL rounds progress, the agent learns to optimize the amount of local training data used by the optimized client. 

The contributions of this paper are summarized as follows:
\begin{itemize}
\item We provide a framework based on DRL to select local training data that a client uses to optimize its performance. Additionally, we investigate and present the results of our proposed framework using well-known FL aggregation algorithms.
\item We design two unique functions for the DRL agent to take actions and adapt them to the existing $\epsilon$-Greedy action selection setup.
\item We design a reward function which takes into account the loss of the local client as well as the amount of data utilized in local training.
\item We conduct theoretical analysis and proof for an upper bound on the performance of the optimized client during the FL rounds.
\end{itemize}

\section{Preliminary}
\subsection{Federated Learning (FL)} 
FL is a distributed learning method that preserves data privacy by training models locally on distributed devices. Instead of sharing actual data with a central server, only local models or local model updates are shared. The server implements an aggregation algorithm to combine the local models into a global model which is disseminated back to the local clients. A typical FL workflow is presented in Figure~\ref{fig:fl_workflow}. Formally, given a set of $K$ total clients, denote the overall datasets as $D=\{D_{1}, D_{2}, ..., D_{K}\}$ from all clients, where each client only leverages its local dataset with $N$ samples $D_{k}:\{x_{n}, y_{n}\}^{N}_{n=1}$. In FedAvg~\cite{mcmahan2017communication}, the FL objective can be written as:
\begin{equation}
\min_{w} f^{*}(w) \overset{\Delta}{=} \frac{1}{K}\sum^{K}_{k=1} f_{k}(w)
\end{equation}
Here $w$ represents the global model parameters and $f_{k}(w): \mathbb{R} \mapsto \mathbb{R}$ is the expected local loss of the client defined as $f_{k}(w) \overset{\Delta}{=} \frac{1}{|D_{k}|} f_{k}(w, D_{k})$ where $f_{k}$ can be substituted for any loss function. The averaging algorithm can also be replaced by other algorithms such as FedMedian~\cite{yin2018byzantine} and FedCDA~\cite{wang2024fedcda}.

\subsection{Reinforcement Learning (RL)} 
\begin{figure}[t]
\centering
    \begin{subfigure}[b]{0.45\textwidth}
        \centering
        \includegraphics[width=0.75\linewidth, trim=100 100 100 100, clip]{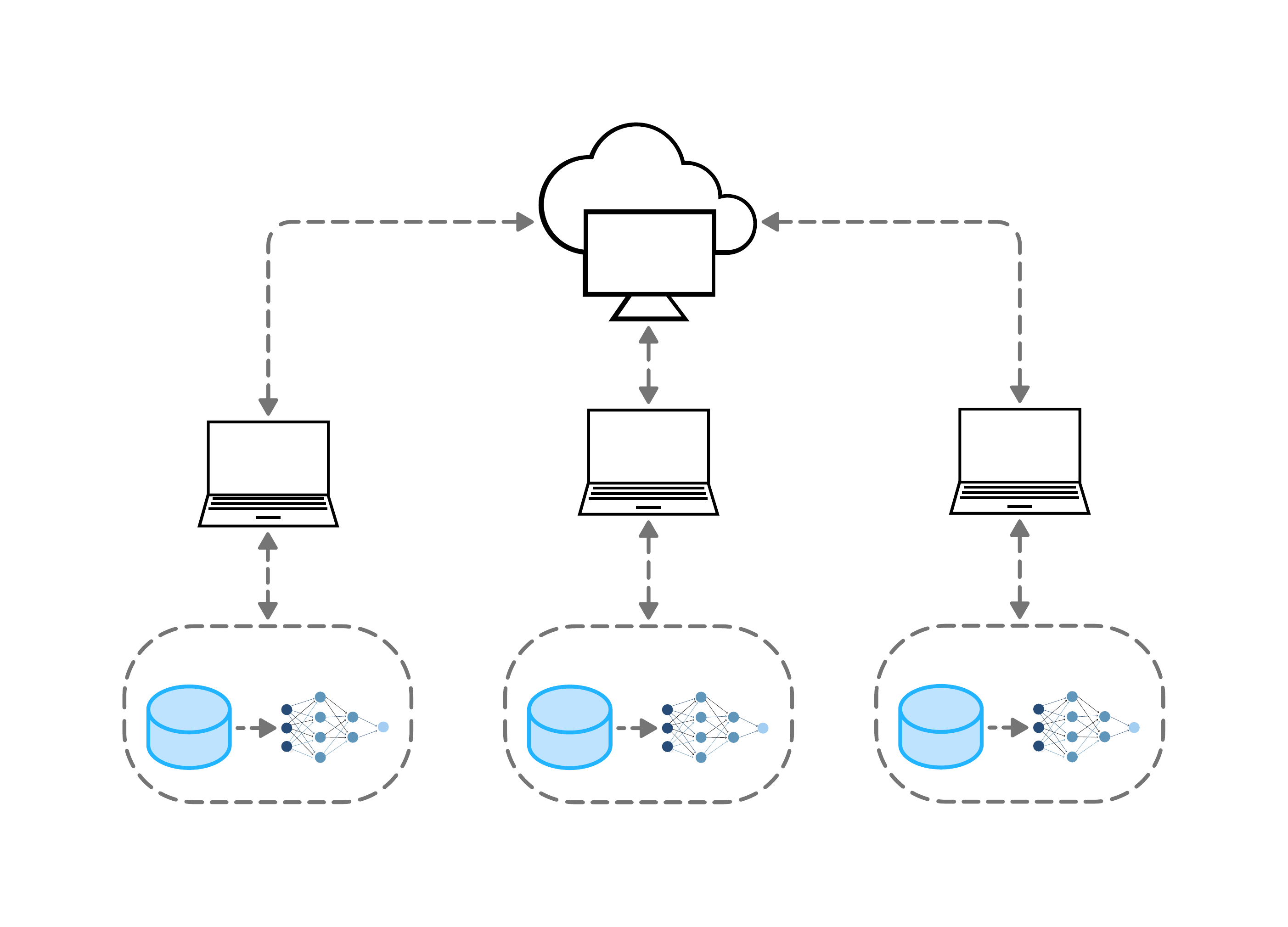}
        \captionsetup{font=footnotesize}
        \caption{FL Workflow. Multiple clients participate in learning a joint model through server aggregation.}
        \label{fig:fl_workflow}
    \end{subfigure}
    \begin{subfigure}[b]{0.45\textwidth}
        \centering
        \includegraphics[width=0.75\linewidth, trim=100 100 100 100, clip]{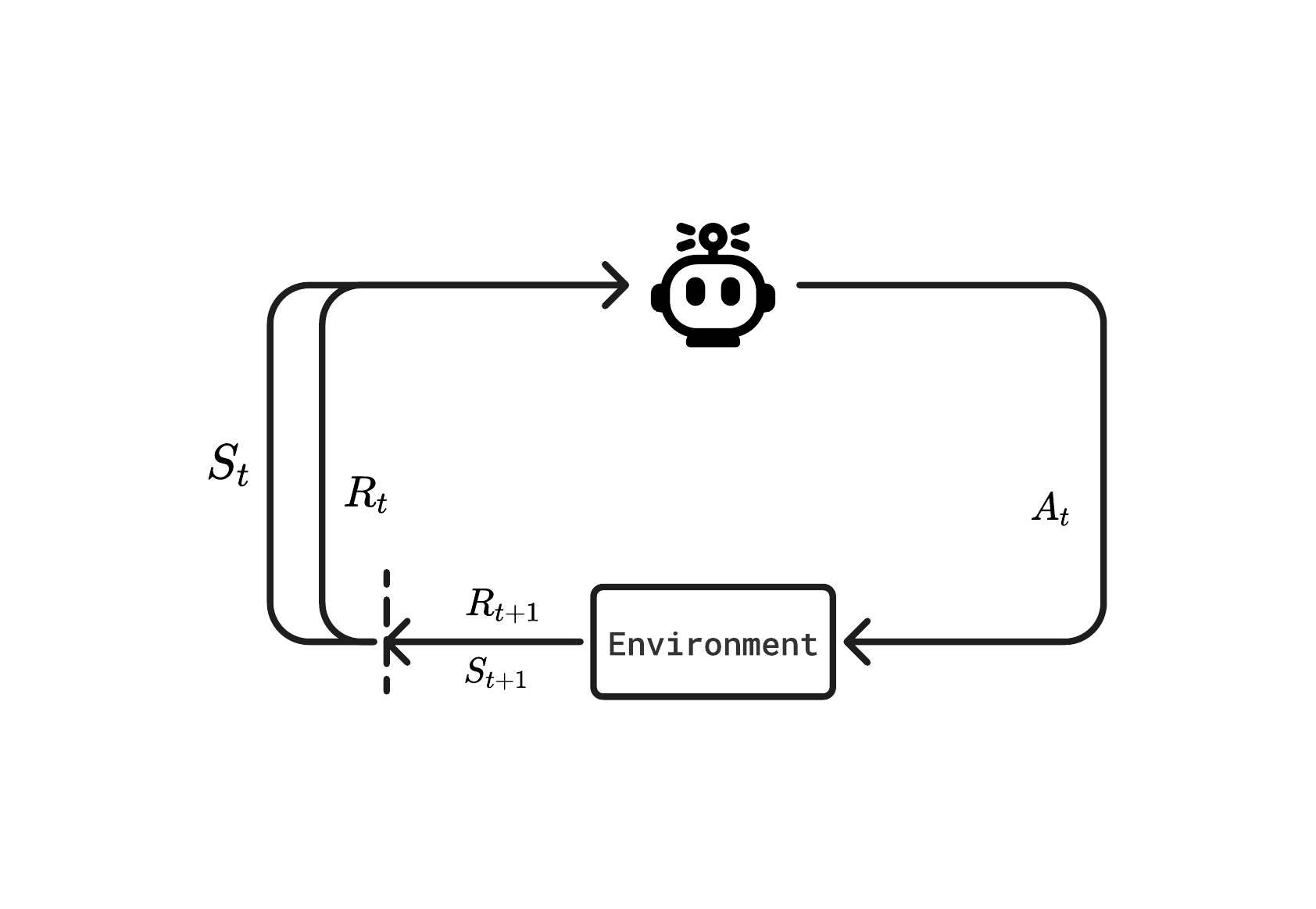}
        \captionsetup{font=footnotesize, justification=raggedright}
        \caption{Agent Environment Interaction. The RL agent receives the initial state of the environment $S_t$ and takes an action $A_t$. This transitions the environment to state $S_{t+1}$ and the RL agent receives the reward $R_{t}$}
        \label{fig:agent_environment}
    \end{subfigure}
    \captionsetup{font=footnotesize, justification=raggedright}
    \caption{FL Workflow \&\ RL Agent Environment Interaction.}
    \vspace{-2mm}
\end{figure}

RL enables building systems in which agents interact with environments to accomplish one or many tasks. Generally, RL systems are modeled as Markov Decision Processes (MDP). At time step $t$, the agent observes an initial state $S_t \in S $ of the environment. Following a policy $\pi_{t}(\cdot|s)$, which maps states to actions, the agent takes an action $A_t \in A$. This transitions the environment to the next state $S_{t+1}$ and the agent receives a reward signal $R_{t+1} \in \mathbb{R}$, informing the agent about the quality of its action. As shown in Figure~\ref{fig:agent_environment}~\cite{sutton2018reinforcement}, the agent environment interaction model gives rise to \textit{trajectories} $(S_{t}, A_{t}, R_{t+1}, S_{t+1}...)$. The expected total reward is given as $G_{t}{=}\sum\limits^{\infty}_{k=0} \gamma^{k} R_{t+k+1}$, where $\gamma \in (0, 1)$ is the discount factor. The value of a given state is meausred by the \textit{State-Value Function} $V_{\pi}(s){=}\mathbb{E}_{\pi}[G_t | S_t = s]$, and the quality of an action paired with a state is given by the \textit{Action-Value Function} $q_{\pi}(s, a) = \mathbb{E}_{\pi}[G_t | S_t = s, A_t = a]$. The RL objective is to find an optimal policy $\pi_{*}$ which maximizes an agent's total return. Policy $\pi_{*}$ shares the optimal state-value function $v_{*}(s)\overset{.}{=}\underset{\pi}{\max} v_{\pi}(s)$ and the optimal action-value function $q_{*}(s, a)\overset{.}{=}\underset{\pi}{\max} q_{\pi}(s, a)$.  \noindent\emph{Deep Reinforcement Learning (DRL)} combines the function approximation ability of Deep Learning (DL) with RL for sequential decision making. This enables RL systems which can generalize to large state and continuous action spaces. Using DL this process is accomplished by mapping large state spaces to features and features to actions. In recent years, DL has been extended to RL methods~\cite{gao2024deep}~\cite{DBLP:journals/corr/LillicrapHPHETS15}~\cite{liu2024integrating}. 

\section{Problem Setup and Framework}
\begin{figure*}[t]
\vspace{-1mm}
    \centering
    \includegraphics[width=1.0\linewidth, height=5.25cm]{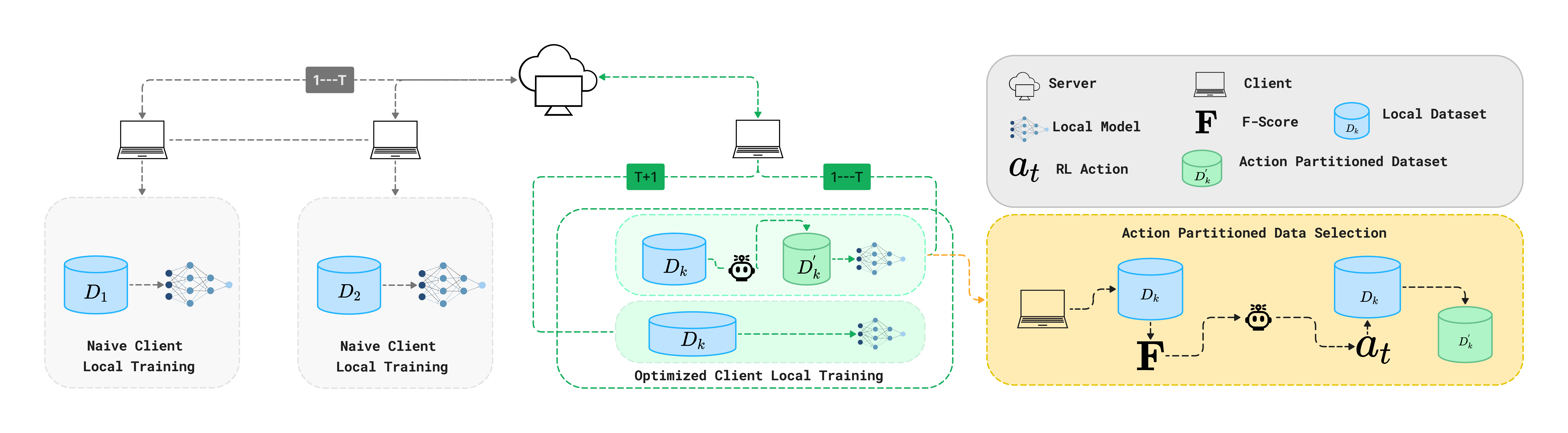}
    \captionsetup{font=footnotesize}
    \caption{Framework for optimized client training and action partitioned data selection.}
    \label{fig:framework}    
\end{figure*}

Given the local dataset ${D_{k}}$ of a client, our target is to optimize the amount of training data $D^{'}_{k}$ for FL. We use the performance of the aggregated server model as the current state $s_{t}$.  The action $a_t$ is defined as a vector that contains the percentage of samples used for training in each class. Using the performance change between the aggregated server model and the local model, we calculate the reward $r_{t}$ with a designed reward function. We train the policy $\pi_{\theta}$ parameterized with $\theta$ based on $r_{t}$, generated from the loss of the aggregated model $\mathcal{L}_{agg}$ and the loss of the client's local model parameters $\mathcal{L}_{l}$ based on local training. The training process encourages $\pi_{\theta}$ to optimize the percentage of data used in each class to create $D^{'}_{k}$ in subsequent FL rounds. After the FL rounds, we leverage the complete dataset ${D_{k}}$ to fine-tune the optimized client until its performance converges. Using ${D_{k}}$ in the final training rounds gives the optimized client an incremental boost in performance resulting from unused data in the previous rounds.  With the proposed framework, we can not only optimize local training but also ensure that the data changes on the optimized client have little impact on the performance of naive clients. Figure~\ref{fig:framework} shows the framework to optimize the percentage of data in each class, by the agent (highlighted in green) to be optimized.

\section{Proposed Method}
Given a total of $T$ FL communication rounds, to train the DRL agent, we use the class-wise performance measured on the local training dataset after server aggregation in communication round $t$ as the current state $s_{t}$. In our experiments, we use the \text{F1}-Score given by $\text{F1} = \frac{2PR}{P + R}$ as a performance measure where $P$ and $R$ are Precision and Recall, respectively~\cite{chinchor1993muc}~\cite{goodfellow2016deep}. Note that \text{F1}-Score can easily be substituted for a different performance measure. Using state $s_{t}$, the policy outputs a vector action $a_{t}$ that contains weights $z_{c}$ for each class $c$ in the local dataset. Formally, for $K$ total clients in the FL setup, with client $k$ as the client to be optimized, $D_{k}: \{X, Y\}$ as the local dataset, and $w_{t}$ as the server aggregated parameters in communication round $t$, the state is given as:
\begin{equation}\label{state}
    s_{t} = \text{F1}(\hat{f}_{k}(X; w_{t}), Y)
\end{equation}
Here, $f_{k}(w_t)$ is the local model of the optimized client parameterized with the server aggregated parameters. Furthermore, we implement two action selection strategies and adapt them to the $\epsilon$-Greedy method. Using policy $\pi_{\theta}$ and user-defined lower/upper bounds as $b_{l}$, $b_{u}$, respectively, the action in round $t$ is given by:
\begin{equation}\label{action}
    \begin{split}
        {a_{t}} =\pi_{\theta}(s_t) = \begin{bmatrix} z_1, z_2, \hdots, z_C \end{bmatrix} \Rightarrow \{z \in \mathbb{R}, {b_{l} \leq z \leq {b_{u}}}\} \hspace{2mm} \\ \text{s.t.} \hspace{2mm} {b_{l}}, {b_{u}} \in (0, 1],
    \end{split}
\end{equation}

{\bf$\epsilon$-Greedy Normalized Action} implements a normalized version of the action generated by the policy. The action vector ${a_{t}}$ is first normalized and multiplied by the total samples in $D_{k}$ to get the count of data samples for each class $c$.
\begin{equation}
    {a'}_t \leftarrow
    \frac{{a_{t}}}{|a_{t}|_1}|D_{k}|
\end{equation}
Here ${a'} = \begin{bmatrix}a'_1, a'_2, \cdotsTwo, a'_C\end{bmatrix}$ is a vector of data sample counts for each class. The class counts are then adjusted to not exceed the total number of samples available for each class. Given $|D_{k_{c}}|$ as the maximum class count for each class in $D_{k}$, the $\epsilon$-Greedy Normalized Action is given as:
\begin{subnumcases}{{a'_{t}} \leftarrow}
    \frac{max(a'_1, |D_{k_{1}}|)}{|D_{k_{1}}|}, \hdots, \frac{max(a'_C, |D_{k_{C}}|)}{|D_{k_{C}}|}& $\text{if } \epsilon$, \label{normalized_action1}\\
    \underset{a}{\argmax}Q(a) & $1 - \epsilon$ \label{normalized_action2}\
\end{subnumcases}

{\bf$\epsilon$-Greedy Weighted Metric Action} implements a look-back period $\eta$, where every $\eta$ communication rounds, the optimized client computes the difference in the absolute value between the current \text{F1}-Score and the \text{F1}-Score from $\eta$ rounds ago. For every class where the \text{F1}-Score has decreased, since $\eta$ rounds, the weight of that class is increased by the difference and then normalized. Formally, 
given $\text{F1}_{t}$ and $\text{F1}_{t-\eta}$ as the \text{F1}-Scores in the current round and the \text{F1}-Scores from $\eta$ rounds ago, and $\Delta \text{F1} = \text{F1}_{t} - \text{F1}_{t-\eta}$ as their difference, the normalized difference is then given as:

\begin{subnumcases}{{|\Delta \text{F1}| \leftarrow}}
  1 + |\Delta \text{F1}| & $\text{if } \Delta \text{F1} < 0$, \\
  1 & $\Delta \text{F1} >= 0$.
  \label{normalized_difference}
\end{subnumcases}

\begin{equation}\label{weighted_metric}
    \hat{\Delta{\text{F1}}} = 
    \frac{|\Delta \text{F1}|} {|\Delta \text{F1}|_1}.
\end{equation}
Using \eqref{weighted_metric} and ensuring that the percentage increase does not exceed the total size of the local dataset, the $\epsilon$-Greedy Weighted Metric Action is given as:

\begin{subnumcases}{{a_{t}} \leftarrow }
    \max{(\hat{\Delta \text{F1}}{a_{t}}, 1)} & $\text{if } \epsilon$, \label{weighted_metric_action1}\\
    \underset{a}{\argmax}Q(a) & $1 - \epsilon$ \label{weighted_metric_action2}\
\end{subnumcases}

The optimized client utilizes the action generated by the DRL agent to create an \emph{Action Partitioned Dataset}, denoted by $D^{'}_{k}$ such that $D^{'}_{k} \subset D_{k}$. Figure~\ref{fig:framework} (highlighted in yellow) illustrates this procedure. As the FL rounds progress we implement a loss estimation mechanism. Specifically, the optimized client waits for $\tau$ communication rounds and then in every subsequent round estimates the local loss for the local training update after server aggregation in the following round. Based on the assumption that, as the FL rounds progress, the client's local training on $D_{k}$ is supposed to produce a lower training loss as we use the following piece-wise reward function:

\begin{subnumcases}{R_{t} \Leftarrow}
    \frac{\mathcal{L}_{\text{agg}} - \mathcal{L}_{l}}{\mathcal{L}_{l}}\frac{1}{\mu_{a_{t}} - \lambda} & $\text{if } T < \tau$, \label{eq:reward1} \\
  \frac{\mathcal{L}_{\text{agg}} - \mathcal{L}_{\text{est}}}{\mathcal{L}_{\text{est}}}\frac{1}{\mu_{a_{t}} - \lambda} & $\text{otherwise}$ \label{eq:reward2} \
\end{subnumcases}

Here, $\mathcal{L}_{\text{agg}}$ is the loss on the local dataset after server aggregation, $\mathcal{L}_{l}$ is the loss on the local dataset after local training, $\mathcal{L}_{\text{est}} = -ue^{-vt}$ is the estimated loss, $\mu_{a_{t}} = \frac{1}{|{a_{t}}|}\sum{a_{t}}$ is the mean action generated by the policy $\pi_{\theta}$, and $\lambda$ is a user defined parameter which normalizes the reward by controlling the amount of local training data generated by the policy. As part of loss estimation, $\mathcal{L}_{\text{est}}$, we fit a non-linear curve \citep{vugrin2007confidence} to estimate the parameters, $u$ and $v$, after each FL round past $\tau$ rounds. Using \eqref{state}, \eqref{normalized_action1}, \eqref{normalized_action2}, \eqref{weighted_metric_action1}, \eqref{weighted_metric_action2}, \eqref{eq:reward1}, and \eqref{eq:reward2} we can generate RL trajectories $(s_{t}, a_{t}, R_{t+1}, s_{t+1}...)$. The actor-critic paradigm, in DRL, then enables learning a parameterized actor policy $\pi_{\theta}(a_{t}|s_{t})$ which outputs an action $a_{t}$ given the current state $s_{t}$, as well as a parameterized critic network $v_{\varphi}(s)$ which approximates a state value function. The critic network can be updated through Mean Squared Error $\nabla\mathcal{L}(\varphi|s_{t}, a_{t}) = (\hat{Q}_{n}(s_{t}, a_{t}) - v_{\varphi}(s_t))^{2}$ followed by the update for policy network $\nabla_{\theta}\mathcal{L}(\theta|s_{t}, a_{t}) = \hat{Q}_{n}(s_{t}, a_{t}).\nabla_{\theta}log \pi_{\theta}(a_{t}|s_{t})$ where $\hat{Q}_{n}(s_{t}, a_{t}) = \sum^{n-1}_{k=0}r_{t+k} + v_{\varphi}(s_{t+n})$ is the n-step target \cite{plaat2022deep}. 

\subsection{Algorithm.} We present the algorithm to train the optimized client both from the server and from the client side. The server-side implementation follows a typical FL set-up, whereas the client-side implementation includes optimized training for the client both during and after the FL rounds. We use DDPG (Deep Deterministic Policy Gradient) \cite{DBLP:journals/corr/LillicrapHPHETS15} to train the RL policy. 


\begin{algorithm}[h]
\caption{\textbf{Optimized Client Training}}
\textbf{Parameters:} $K$(number of total clients), $C \in (0,1) \mapsto \mathbb{R}$ (predetermined ratio of clients to participate in each round), $FederatedAggregation$ (federated learning aggregation algorithm.), $E$ (local train epochs) \\
\label{alg: optimized_client_training}
\textbf{Server:}
\begin{algorithmic}[1]
    \State{initialize $w_{0}$} 
    \For{ round $t = 0, 1, 2, \cdots, T$}
        \State {$S_{t} = \{\text{random sample of $C*K$ Clients}$}\}
        \For{$k \in S$ \textbf{in parallel}} 
        \State {$w_{t+1} = OptimalClientTrain(k, w_{t})$}
        \EndFor
    \State{$w_{t+1} = FederatedAggregation(S_{t})$}
\EndFor 
\end{algorithmic}
\textbf{Client: \Comment{$OptimalClientTrain(k, w_{t})$}}
\begin{algorithmic}[1]
\While{$t \leq T$}
    \State{Compute $s_{t}$ using Equation. \ref{state}}
    \State{Compute $a_{t}$ using Equation. \ref{action}}
    \State{$D^{'}_{k} \leftarrow D_{k}(a_{t})$ \Comment{$ActionPartitionedDataset$}}
    \State{$B \leftarrow \{\text{Create batches of size } B \in D^{'}_{k}\}$}
    \For{$e = 1, 2, 3 \cdots \textbf{ in } E$}
        \For{$b \textbf{ in } B$}
            \State{$w_{t} \leftarrow w_{t} - \eta\nabla l(w; b)$}
        \EndFor
    \EndFor
    \State{\text{return $w_{t}$ to server}}
\EndWhile
\State{$B \leftarrow \{\text{Create batches of size } B \in D_{k}\}$}
\For{$e = 1, 2, 3 \cdots \textbf{ in } E$}\Comment{$Until Convergence$}
        \For{$b \textbf{ in } B$}
            \State{$w_{t} \leftarrow w_{t} - \eta\nabla l(w; b)$}
        \EndFor
\EndFor
\end{algorithmic}
\end{algorithm}

\subsection{Analysis.} In this section, we investigate whether the performance of the optimized client has an upper bound during the FL rounds. Based on the assumption that using more data leads to higher performance, we note that the performance of the optimized client will not be as good as if it were trained on its entire local dataset. Under this assumption, we elucidate the answer to one main question: \textit{Is there a performance upper bound for the optimized client during the FL rounds?} 

\begin{proposition}
  Given $s_{t}$ and  $a_{t} = \begin{bmatrix} z_1, z_2, \hdots, z_C \end{bmatrix} \Rightarrow \{z \in \mathbb{R}, {0 \leq z \leq 1}\}$ as the state and the action taken by the policy, let $z$ be the radius of a unit circle representing the total available sample size for each class in the \textit{Action Partitioned Dataset} $D^{'}_{k_{1,2,3 \cdots C}} \forall c \in C$. Let $A = \begin{bmatrix} Z_1, Z_2, \hdots, Z_C \end{bmatrix} \Rightarrow \{Z \in \mathbb{R}, {0 \leq Z \leq 1}\}$ be a vector representing the total samples for each class in the complete local client dataset $D_{k_{1,2,3 \cdots C}} \forall c \in C$. Additionally, let $\omega = Z_{C}^{2} - z_{c}^{2}$ be the difference in the squared radii. The performance bound, of the client trained on the complete dataset, for class $c$ is defined as the area of the circle:
\begin{equation}\label{client_performance_bound}
    P_{k_{c}} =\pi Z^{2}_{c} 
\end{equation}  
\end{proposition}

\noindent Using \eqref{client_performance_bound}, the total performance of client $k$, on $D_{k}$, is:
\begin{equation}\label{total_client_performance_bound}
    P_{k} =\pi Z^{2}_{1} +\pi Z^{2}_{2} +\pi Z^{2}_{3} +\cdotsTwo+ \pi Z^{2}_{c} 
\end{equation}

\noindent Similarly, using \eqref{client_performance_bound}, the performance of client $k$ on $D^{'}_{k}$ is:
\begin{equation}\label{action_partitioned_client_performance_bound}
    P^{'}_{k} =\pi z^{2}_{1} +\pi z^{2}_{2} +\pi z^{2}_{3} +\cdotsTwo+ \pi z^{2}_{c} 
\end{equation}

\begin{theorem}[Performance Upper Bound]: A client trained on $D^{'}_{k}$ relative to $D_{k}$ has a performance bound given by:
\begin{equation}
    P_{k} - P^{'}_{k} \leq \Omega
\end{equation}
\end{theorem}

\begin{IEEEproof}
\begin{align*}
P_{k} - P^{'}_{k} &= 
\pi Z^{2}_{1} +\pi Z^{2}_{2} +\cdotsTwo+ \pi Z^{2}_{c} - \pi z^{2}_{1} -\pi z^{2}_{2} -\cdotsTwo- \pi z^{2}_{c}  \\
    &= \pi Z^{2}_{1} - \pi z^{2}_{1} + \pi Z^{2}_{2} - \pi z^{2}_{2} +\cdotsTwo+ \pi Z^{2}_{C} - \pi z^{2}_{C} \\
    &= \pi(Z_{1}^{2} - z_{1}^{2})+\pi(Z_{3}^{2} - z_{3}^{2}) +\cdotsTwo+ \pi(Z_{C}^{2} - z_{C}^{2}) \\ 
    &= \pi \omega_{1} +\pi \omega_{2} + \pi \omega_{3} +\cdotsTwo+ \pi \omega_{C} \\
    &\leq \pi \sum^{C}_{c=1} \omega_{c} = \Omega 
\end{align*}  
\end{IEEEproof}
The proof above shows that constructing a dataset $D^{'}_{k}$ by minimizing the difference between the action taken by the policy and the total sample size for each class in $D_{k}$ will lead to better performance by the optimized client during the FL rounds. However, we note that this also presents a trade-off between the performance improvement that the optimized client will benefit from by retaining these additional data samples for training after the FL rounds.

\vspace{-2mm}
\section{Experiment}
\subsection{Setup} We conduct experiments of our proposed methodology using 5 FL baseline aggregation algorithms. These algorithms include FedCDA \cite{wang2024fedcda}, FedProx \cite{li2020federated}, FedMedian \cite{yin2018byzantine}, FedAvgM \cite{hsu2019measuring}, and FedAvg \cite{mcmahan2017communication}. For DRL we use Deep Deterministic Policy Gradients~\cite{lapan2020deep}~\cite{DBLP:journals/corr/LillicrapHPHETS15} and ResNet50 \cite{he2016deep} as the server and the client models. Our experiments are implemented in Python. Additionaly, we use scientific programming libraries including scikit-learn \cite{sklearn_api}, Numpy \cite{harris2020array}, Flower \cite{beutel2020flower}, Scipy \cite{2020SciPy-NMeth}, and PyTorch~\cite{paszke2019pytorch}. All plots are generated using Matplotlib~\cite{Hunter:2007} and Plotly~\cite{plotly}. The experiments are conducted using 3 NVIDIA GeForce RTX 3080 GPUs.

\subsection{Datasets.} We conduct our experiments using 3 benchmark datasets, CIFAR 10, CIFAR 100 \cite{krizhevsky2009learning}, and FashionMNIST \cite{xiao2017/online}. Each dataset contains 10, 100, and 10 classes, respectively. Furthermore, we create non-IID partitions using the Dirichlet Partitioner \cite{yurochkin2019bayesian} for each client as its own local dataset. Each partition is split using 80/20 split, where 80\% of the data is used for training and 20\% of the data is used for validation.

\subsection{Results and Analysis} 
Our experiments show the utility of our proposed method compared to well-established baselines. We conduct 100 FL rounds for 8 clients where each client is trained for 1 local epoch. Through our experiments, we demonstrate the generalization capability of our method in different FL settings. The results from our experiments are summarized in Table \ref{experimental_results}, where we show a comparison of the best mean performance of the naive clients, including Precision, Recall, and Accuracy, relative to the optimized client on the validation datasets. Each two-row combination shows a comparison of the mean performance achieved by all naive clients in the FL setup relative to the best performance of the optimized client, after training on the complete local dataset, $D_{k}$, after the FL rounds.

\begin{figure*}[h]
    \centering
    \captionsetup{font=footnotesize}
    \begin{subfigure}[b]{1.0\textwidth}
    \includegraphics[width=1.0\linewidth, trim={3cm 0 2cm 0.2cm}, clip]{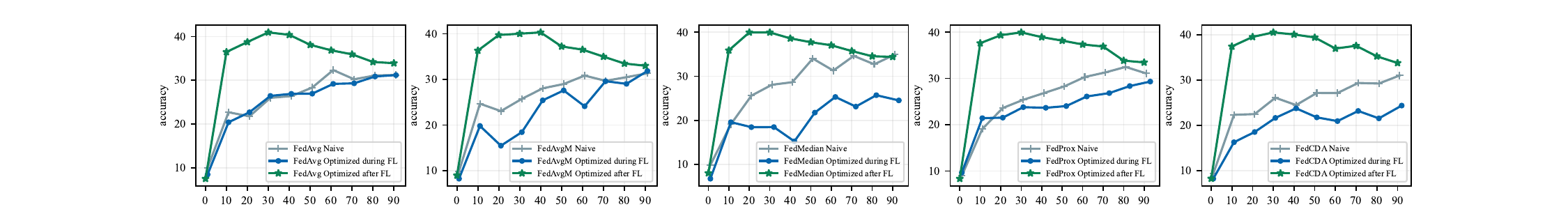}
    \end{subfigure}
    \begin{subfigure}[b]{1.0\textwidth}
    \includegraphics[width=1.0\linewidth, trim={3cm 0 2cm 0.2cm}, clip]{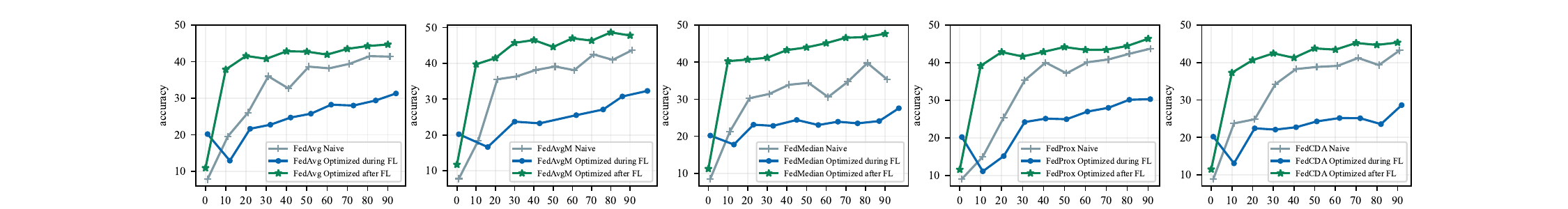}
    \end{subfigure}
    \begin{subfigure}[b]{1.0\textwidth}
    \includegraphics[width=1.0\linewidth, trim={3cm 0 2cm 0.2cm}, clip]{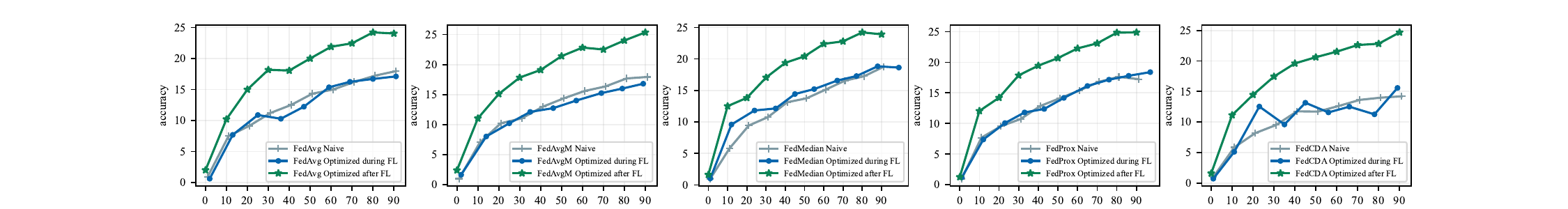}
    \end{subfigure}
    \caption{Mean accuracy in FL rounds. The grey line shows the mean accuracy of all naive clients, whereas the blue line shows the accuracy of the optimized client during the FL rounds. The dark green line shows the accuracy of the optimized client in each epoch after FL rounds.}
    \label{fig:performance_plots}
\end{figure*}

\begin{figure}[t]
    \centering
    \begin{subfigure}[b]{0.5\textwidth}
        \centering
        \includegraphics[width=.75\linewidth, trim=5 5 -50 0, clip]{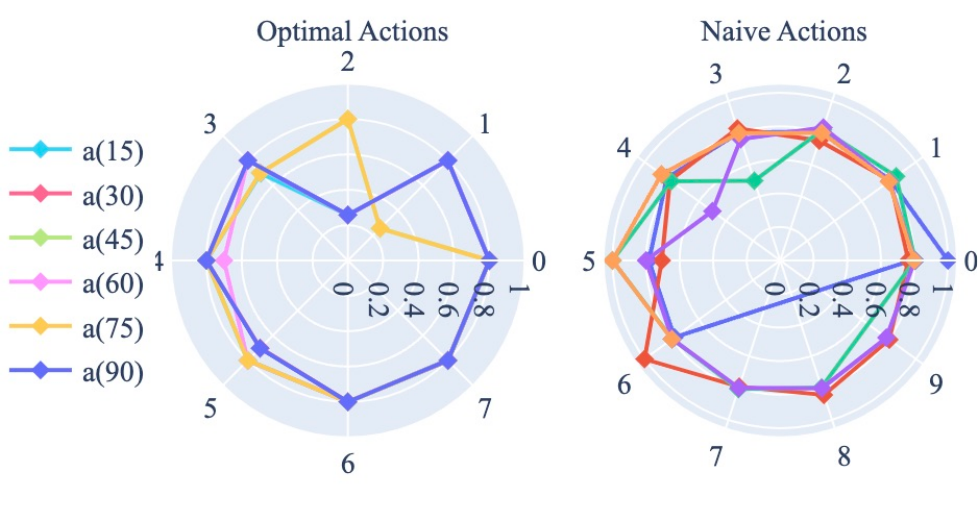}
        \captionsetup{font=footnotesize}
        \caption{Normalized vs. Naive Actions}
        \label{fig:normalized_vs_naive_actions}
    \end{subfigure}
    \begin{subfigure}[b]{0.5\textwidth}
        \centering
        \includegraphics[width=0.75\linewidth, trim=5 5 -50 0, clip]{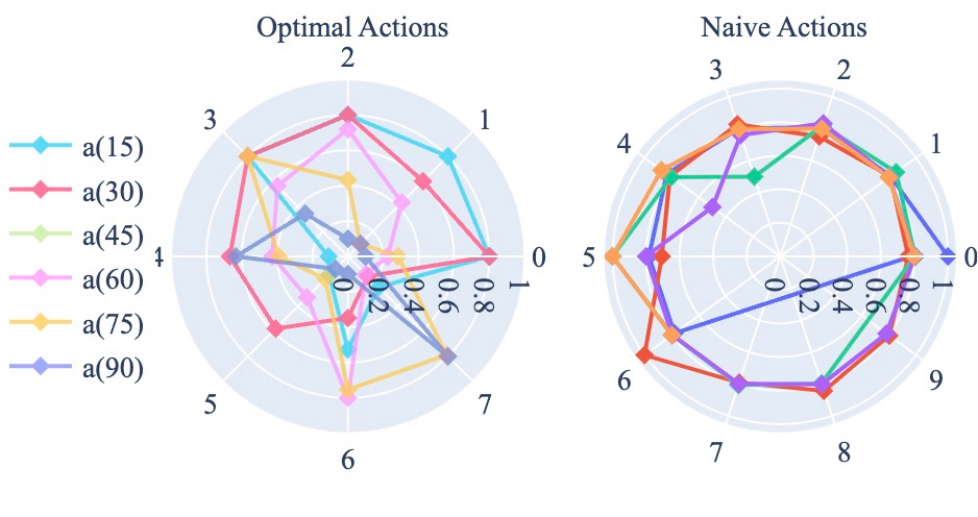}
        \captionsetup{font=footnotesize}
        \caption{Weighted Metric vs. Naive Actions}
        \label{fig:weighted_metric_vs_naive_actions}
    \end{subfigure}
    \captionsetup{font=footnotesize}
    \caption{Optimal actions taken by the RL agent in different federated learning rounds, versus Naive Actions taken by each client.}
    \vspace{-3mm}
    \label{fig:normalized_and_weighted_metric_actions}
\end{figure}

Figure~\ref{fig:performance_plots} shows the mean validation accuracy of the naive clients relative to the optimized client, after each server aggregation, during the FL rounds. The final validation accuracy of the optimized client is plotted as a separate line which shows the best validation accuracy achieved by the optimized client by training on its entire local dataset $D_{k}$ after the FL rounds. It can be observed that the optimized client produces lower performance relative to the naive client during the FL rounds. This phenomenon is illustrated in Figure~\ref{fig:normalized_and_weighted_metric_actions} and attributed to the fact that during FL rounds the optimized client utilizes a smaller proportion of the local dataset relative to all other clients. Figure~\ref{fig:normalized_vs_naive_actions} shows normalized actions and Figure~\ref{fig:weighted_metric_vs_naive_actions} shows weighted metric actions, taken by the DRL agent compared to naive data selection using 80/20 train test split. During the FL phase, the DRL policy determines the minimum viable amount of data necessary for local training. However, after the FL rounds finish, the optimized client is trained on its entire local dataset until it converges. During this phase of local training, the optimized client exhibits superior performance. In addition to the performance improvement of the optimized client after FL rounds, we also observe a considerable increase in convergence speed which can be ascribed to the fact that the optimized client resumes local training using the server aggregated parameters from the final FL round. 


\definecolor{lightgray}{gray}{1.0}
\begin{table*}[t]
\captionsetup{justification=centering}
\caption{\MakeUppercase{Performance comparison with baseline methods.}}
\begin{center}
\renewcommand{\arraystretch}{2.0}
\resizebox{2.05\columnwidth}{!}{\fontsize{12}{10}\selectfont%
\begin{tabular}{c|ccc|ccc|ccc|}
\cline{2-10}
\multirow{1}{*}{} 
& \multicolumn{3}{c|}{\bf Cifar 10} & \multicolumn{3}{c|}{\bf FashionMNIST} & \multicolumn{3}{c|}{\bf CIFAR 100}  \\ \cline{2-10} 
& \multicolumn{1}{c|}{Precision} & \multicolumn{1}{c|}{Recall} &  \multicolumn{1}{c|}{Accuracy} & \multicolumn{1}{c|}{Precision} & \multicolumn{1}{c|}{Recall} & \multicolumn{1}{c|}{Accuracy} & \multicolumn{1}{c|}{Precision} & \multicolumn{1}{c|}{Recall} & \multicolumn{1}{c|}{Accuracy} \\ \hline
\multicolumn{1}{|r|}{FedAvg} & \multicolumn{1}{c|}{$43.07\% \pm 0.46$} & \multicolumn{1}{c|}{$34.00\% \pm 0.64$} & \multicolumn{1}{c|}{$26.64\% \pm 0.52$} & \multicolumn{1}{c|}{$43.90\% \pm 0.49$} & \multicolumn{1}{c|}{$38.55\% \pm 0.62$} & \multicolumn{1}{c|}{$33.97\% \pm 1.01$} & \multicolumn{1}{c|}{$14.96\% \pm 0.83$} & \multicolumn{1}{c|}{$14.18\% \pm 1.50$} & \multicolumn{1}{c|}{$13.41\% \pm 0.37$} \\ \hline
\multicolumn{1}{|r|}{FedAvg + Our Method} & \multicolumn{1}{c|}{$\bf55.35\% \pm \bf2.31$} & \multicolumn{1}{c|}{$\bf42.27\% \pm \bf2.20$} & \multicolumn{1}{c|}{$\bf43.01\% \pm \bf0.27$} & \multicolumn{1}{c|}{$\bf63.55\% \pm \bf0.74 $} & \multicolumn{1}{c|}{$\bf54.61\% \pm \bf1.14$} & \multicolumn{1}{c|}{$\bf50.82\% \pm 0.01$} & \multicolumn{1}{c|}{$\bf20.59\% \pm \bf0.20$} & \multicolumn{1}{c|}{$\bf19.35\% \pm \bf0.69$} & \multicolumn{1}{c|}{$\bf26.40\% \pm \bf0.31$} \\ \hline \hline
\multicolumn{1}{|r|}{FedAvgM} & \multicolumn{1}{c|}{$41.44\% \pm 1.14$} & \multicolumn{1}{c|}{$33.18\% \pm 0.73$} & \multicolumn{1}{c|}{$27.00\% \pm 1.85$} & \multicolumn{1}{c|}{$44.45\% \pm 2.87$} & \multicolumn{1}{c|}{$38.92\% \pm 2.07$} & \multicolumn{1}{c|}{$34.46\% \pm 0.50$} & \multicolumn{1}{c|}{$15.80\% \pm 1.27$} & \multicolumn{1}{c|}{$14.72\% \pm 1.13$} & \multicolumn{1}{c|}{$13.55\% \pm 0.18$} \\ \hline
\multicolumn{1}{|r|}{FedAvgM + Our Method} & \multicolumn{1}{c|}{$\bf55.18\% \pm \bf2.18$} & \multicolumn{1}{c|}{$\bf40.78\% \pm \bf1.65$} & \multicolumn{1}{c|}{$\bf42.57\% \pm 0.36$} & \multicolumn{1}{c|}{$\bf63.41\% \pm \bf1.70$} & \multicolumn{1}{c|}{$\bf55.24\% \pm \bf0.45$} & \multicolumn{1}{c|}{$\bf53.89\% \pm 0.52$} & \multicolumn{1}{c|}{$\bf21.37\% \pm \bf0.73$} & \multicolumn{1}{c|}{$\bf19.30\% \pm \bf1.10$} & \multicolumn{1}{c|}{$\bf26.65\% \pm \bf0.24$} \\ \hline \hline
\multicolumn{1}{|r|}{FedMedian} & \multicolumn{1}{c|}{$35.03\% \pm 1.42$} & \multicolumn{1}{c|}{$29.80\% \pm 3.22$} & \multicolumn{1}{c|}{$28.27\% \pm 2.24$} & \multicolumn{1}{c|}{$33.46\% \pm 0.24$} & \multicolumn{1}{c|}{$33.59\% \pm 0.10$} & \multicolumn{1}{c|}{$32.40\% \pm 0.72$} & \multicolumn{1}{c|}{$14.07\% \pm 1.20$} & \multicolumn{1}{c|}{$13.63\% \pm 1.25$} & \multicolumn{1}{c|}{$13.03\% \pm 0.06$} \\ \hline
\multicolumn{1}{|r|}{FedMedian + Our Method} & \multicolumn{1}{c|}{$\bf58.08\% \pm \bf2.36$} & \multicolumn{1}{c|}{$\bf42.64\% \pm \bf1.37$} & \multicolumn{1}{c|}{$\bf43.24\% \pm 0.32$} & \multicolumn{1}{c|}{$\bf67.76\% \pm \bf2.55$} & \multicolumn{1}{c|}{$\bf58.90\% \pm 2.52$} & \multicolumn{1}{c|}{$\bf53.74\% \pm 0.70$} & \multicolumn{1}{c|}{$\bf21.80\% \pm \bf0.61$} & \multicolumn{1}{c|}{$\bf19.95\% \pm \bf0.58$} & \multicolumn{1}{c|}{$\bf26.40\% \pm \bf0.03$} \\ \hline \hline
\multicolumn{1}{|r|}{FedProx} & \multicolumn{1}{c|}{$41.90\% \pm 0.02$} & \multicolumn{1}{c|}{$33.28\% \pm 1.28$} & \multicolumn{1}{c|}{$27.60\% \pm 0.15$} & \multicolumn{1}{c|}{$43.60\% \pm 1.17$} & \multicolumn{1}{c|}{$39.18\% \pm 1.20$} & \multicolumn{1}{c|}{$33.58\% \pm 0.94$} & \multicolumn{1}{c|}{$7.53\% \pm 0.38$} & \multicolumn{1}{c|}{$8.54\% \pm 0.30$} & \multicolumn{1}{c|}{$6.76\% \pm 0.32$} \\ \hline
\multicolumn{1}{|r|}{FedProx + Our Method} & \multicolumn{1}{c|}{$\bf55.82\% \pm \bf1.11$} & \multicolumn{1}{c|}{$\bf44.71\% \pm \bf1.41$} & \multicolumn{1}{c|}{$\bf43.36\% \pm \bf0.87$} & \multicolumn{1}{c|}{$\bf62.36\% \pm \bf1.12$} & \multicolumn{1}{c|}{$\bf55.17\% \pm \bf1.56$} & \multicolumn{1}{c|}{$\bf50.13\% \pm \bf0.25$} & \multicolumn{1}{c|}{$\bf22.14\% \pm \bf1.77$} & \multicolumn{1}{c|}{$\bf20.48\% \pm \bf1.01$} & \multicolumn{1}{c|}{$\bf28.87\% \pm \bf0.48$}  \\ \hline \hline
\multicolumn{1}{|r|}{FedCDA} & \multicolumn{1}{c|}{$37.66\% \pm 2.32$} & \multicolumn{1}{c|}{$28.08\% \pm 2.31$} & \multicolumn{1}{c|}{$22.97\% \pm 0.57$} & \multicolumn{1}{c|}{$46.11\% \pm 0.15$} & \multicolumn{1}{c|}{$39.42\% \pm 0.14$} & \multicolumn{1}{c|}{$33.00\% \pm 1.69$} & \multicolumn{1}{c|}{$7.10\% \pm 0.64$} & \multicolumn{1}{c|}{$6.94\% \pm 0.60$} & \multicolumn{1}{c|}{$6.31\% \pm 0.31$} \\ \hline
\multicolumn{1}{|r|}{FedCDA + Our Method} & \multicolumn{1}{c|}{$\bf54.79\% \pm \bf2.66$} & \multicolumn{1}{c|}{$\bf45.60\% \pm \bf0.02$} & \multicolumn{1}{c|}{$\bf40.43\% \pm \bf1.41$} & \multicolumn{1}{c|}{$\bf65.32\% \pm 3.13$} & \multicolumn{1}{c|}{$\bf57.91 \pm \bf4.59$} & \multicolumn{1}{c|}{$\bf48.00\% \pm \bf2.05$} & \multicolumn{1}{c|}{$\bf23.12\% \pm \bf0.99$} & \multicolumn{1}{c|}{$\bf20.82\% \pm 0.62$} & \multicolumn{1}{c|}{$\bf30.54\% \pm \bf0.25$} \\ \hline
\end{tabular}%
}
\label{experimental_results}
\end{center}
\end{table*}

\begin{table}[t]
\definecolor{lightgray}{gray}{1.25}
\captionsetup{justification=centering}
\caption{\MakeUppercase{Ablation Study}}
\begin{center}
\renewcommand{\arraystretch}{1.25}
\resizebox{1.\columnwidth}{!}{\fontsize{10}{11}\selectfont%
\begin{tabular}{r|c|c|c|}
\cline{2-4}
\multirow{1}{*}{} 
& \multicolumn{1}{|c|}{Precision} & \multicolumn{1}{|c|}{Recall} &  \multicolumn{1}{|c|}{Accuracy} \\ \hline 
\multicolumn{1}{|r|}{FedAvg (original)} & \multicolumn{1}{|c|}{$46.92\% \pm 1.87$} & \multicolumn{1}{|c|}{$40.06\% \pm 1.26$} & \multicolumn{1}{|c|}{$34.98\% \pm 1.17$} \\ \hline 
\multicolumn{1}{|r|}{FedAvg (with optimized client)} & \multicolumn{1}{|c|}{$50.44\% \pm 0.31$} & \multicolumn{1}{|c|}{$47.48\% \pm 0.95$} & \multicolumn{1}{|c|}{$26.91\% \pm 0.65$} \\ \hline \hline
\multicolumn{1}{|r|}{FedAvgM (original)} & \multicolumn{1}{|c|}{$41.42\% \pm 3.39$} & \multicolumn{1}{|c|}{$31.31\% \pm 3.27$} & \multicolumn{1}{|c|}{$30.65\% \pm 3.13$} \\ \hline
\multicolumn{1}{|r|}{FedAvgM (with optimized client)} & \multicolumn{1}{|c|}{$43.72\% \pm 4.87$} & \multicolumn{1}{|c|}{$41.95\% \pm 4.84$} & \multicolumn{1}{|c|}{$23.19\% \pm 2.14$} \\ \hline \hline
\multicolumn{1}{|r|}{FedMedian (original)} & \multicolumn{1}{|c|}{$34.46\% \pm 1.86$} & \multicolumn{1}{|c|}{$39.07\% \pm 4.99$} &  \multicolumn{1}{|c|}{$33.64\% \pm 1.54$}\\ \hline
\multicolumn{1}{|r|}{FedMedian (with optimized client)} & \multicolumn{1}{|c|}{$35.81\% \pm 0.31$} & \multicolumn{1}{|c|}{$42.27\% \pm 1.53$} & \multicolumn{1}{|c|}{$22.95\% \pm 0.24$} \\ \hline \hline
\multicolumn{1}{|r|}{FedProx (original)} & \multicolumn{1}{|c|}{$44.18\% \pm 1.84$} & \multicolumn{1}{|c|}{$38.31\% \pm 1.24$} & \multicolumn{1}{|c|}{$33.85\% \pm 0.61$} \\ \hline
\multicolumn{1}{|r|}{FedProx (with optimized client)} & \multicolumn{1}{|c|}{$47.73\% \pm 4.01$} & \multicolumn{1}{|c|}{$46.24\% \pm 3.34$} & \multicolumn{1}{|c|}{$25.22\% \pm 0.93$} \\ \hline \hline
\multicolumn{1}{|r|}{FedCDA (original)} & \multicolumn{1}{|c|}{$46.65\% \pm 1.55$} & \multicolumn{1}{|c|}{$39.42\% \pm 2.67$} & \multicolumn{1}{|c|}{$31.14\% \pm 2.39$} \\ \hline
\multicolumn{1}{|r|}{FedCDA (with optimized client)} & \multicolumn{1}{|c|}{$50.91\% \pm 0.69$} & \multicolumn{1}{|c|}{$48.31\% \pm 1.53$} & \multicolumn{1}{|c|}{$24.48\% \pm 0.23$} \\ \hline 
\end{tabular}%
}
\end{center}
\label{table:ablation_results}
\vspace{-10mm}
\end{table}

\subsection{Ablation Study.}  To further verify the effectiveness of the proposed method, we conduct an ablation study. Specifically, we use naive actions for every client. Naive actions correspond to each client's dataset being split using the 80/20 split. The results of our ablation experiments are summarized in Table~\ref{table:ablation_results}. It is evident from the results that learning a DRL policy to partition the local dataset, during the FL rounds, followed by training on the complete local dataset, yields improved overall performance for the optimized client. We note that compared to naive actions, optimized actions can calculate the amount of useful data and utilizing more data than necessary could result in client models offsetting each other's local updates during model aggregation.

\subsection{Discussion.} Our experimental findings show that training a client on a partition of its own local data allows the client to improve its performance during the FL rounds, and benefit considerably by training on the complete dataset after the FL rounds. Through DRL, a parameterized policy can be learned and optimized on the client's local performance as the client interacts with the server, enabling the client to dynamically create partitions of its local training data. During FL, the optimized client benefits from aggregation, whereas after FL the optimized client leverages information from unused samples to further improve its performance. We note that training on a smaller partition of the local data can cause the optimized client to marginally lag in performance relative to the naive clients during the FL rounds. This motivates us to further investigate potential solutions that can maintain competitive performance during the FL rounds.

\section{Related Works.} Since our work prioritizes improving client performance in a FL setting, we provide an overview of related methods and techniques that address data heterogeneity issues and improve client personalization. These areas form the cross-section of technologies that enable our research. 

\subsection{Data Heterogeneity Issues.} Data Heterogeneity can potentially have an adverse impact on model convergence and final model performance~\cite{kim2023depthfl}~\cite{10651397}~\cite{liu2025towards}~\cite{yu2023turning}. To address this issue, since FedAVG \cite{mcmahan2017communication}, many variants of FL aggregation algorithms have been proposed. FedProx \cite{li2020federated} add a proximal term to bring the local models closer to the global model. FedDC~\cite{gao2022feddc} addresses data heterogeneity through a local drift variable which improves model consistency and performance, resulting in faster convergence across diverse tasks. FedCDA~\cite{wang2024fedcda} addresses this issue in a cross-round setting by selecting and aggregating local models that minimize divergence from the global model. Tang et al.~\cite{tang2024fedimpro} improve client updates in an attempt to improve global model performance and \cite{huang2024stochastic} introduce two compressed FL algorithms that achieve improved performance under arbitrary data heterogeneity. Li et al. and Wang et al.~\cite{li2024fedcompass}~\cite{wang2024tackling} study data heterogeneity in an asynchronous setting and propose methods for caching local client updates to measure each client's contribution to the global model as well as to reduce the staleness of clients in global model updates. 

\subsection{Personalization and Optimization. } Due to device and data heterogeneity, training client models on local data can result in better outcomes relative to participating in FL~\citep{wu2020personalized}. Personalization attempts to improve client performance by accounting for the local data distribution of a client~\cite{jiang2024heterogeneous}~\cite{10650714}~\cite{10650798}. Huang et al.~\cite{huang2021personalized} propose a method, FedAMP, which enables a message passing mechanism among similar clients to improve performance between them. FedALA \cite{zhang2023fedala} achieves better personalization by adapting to the local objective through element-wise aggregation of global and local models. pFedBEA~\cite{10650954} and FedPAC~\cite{scott2024pefll} implement the sharing of client models and a regularization term, respectively, sharing common attributes and accounting for the label distribution shift among clients. This enables learning shared feature extraction layers in deep neural networks and shared classification heads. Cheng et al. and Wang et al.~\cite{cheng2024momentum}~\cite{wang2024fedhyper} study momentum and hyperparameter optimization to gain faster convergence. Chanda et al. \cite{chanda2024bayesian} strive for better performance by training clients on coresets of their local training data, assigning a weight vector to each client, which acts as the coreset weight, while~\citep{10630939} study personalization for client fairness.

\section{Conclusion}
In our work, we propose a novel method to train clients for improved personalization through efficient usage of the client's local data. We leverage the planning and the sequential decision making capabilities of DRL. Our method shows that efficient utilization of local data enables a client to have better performance compared to naive training during FL. Additionally, we show that a learned DRL policy, by designing an adequate reward function, can help the client optimize its performance. We note that utilizing a smaller partition of local data can result in lower performance during the FL rounds and to remedy this we establish a theoretical upper bound on the client performance, and present a trade-off between improving performance during FL rounds versus improving performance after FL. Overall, we hope that our work encourages more research in utilizing DRL to orchestrate client training in FL and future works extend the ideas presented in our work to multiple clients using multi-agent and model-based RL systems.

\bibliographystyle{IEEEtranS}
\bibliography{bibliography}

\end{document}